\documentclass[10pt,twocolumn,letterpaper]{article}

\usepackage[pagenumbers]{cvpr} 

\usepackage{cuted} 

\usepackage{currfile} 

\usepackage{caption} 
\definecolor{cvprblue}{rgb}{0.21,0.49,0.74}
\usepackage[pagebackref,breaklinks,colorlinks,allcolors=cvprblue]{hyperref}
\usepackage{algorithm}
\usepackage{bm}
\usepackage{algpseudocode}
\usepackage{booktabs}
\usepackage{multirow}
\usepackage{amssymb}
\usepackage{xcolor}
\usepackage{colortbl}

\definecolor{oursblue}{rgb}{0.88,0.95,0.99}

\def\confName{CVPR}
\def\confYear{2027}

\title{Verify then Steer: Progress Hallucination-Aware Recovery for Continuous Vision-and-Language Navigation}

\author{Jiewen Hou$^{1,2}$ \quad Guanxiang Qi$^{2}$ \quad Meina Kan$^{1,2}$ \quad Shiguang Shan$^{1,2}$ \quad Xilin Chen$^{1,2}$\\
$^{1}$Institute of Computing Technology, Chinese Academy of Sciences\\
$^{2}$University of Chinese Academy of Sciences\\
{\tt\small houjiewen24z@ict.ac.cn}
}

\begin{document}
\maketitle

\begin{abstract}
Vision-and-Language Navigation (VLN) requires an agent to follow language instructions to reach a target in an unseen 3D environment. However, existing methods still struggle to recover from route deviations at test time. We investigate this limitation through the lens of instruction progress, and trace it to a phenomenon we term \textbf{progress hallucination}: agents misinterpret off-route observations as evidence of task progress, causing their internal instruction progress to advance beyond the actual task stage and hindering recovery based on earlier sub-instructions. To address this failure mode, we propose PHAR (\textbf{P}rogress \textbf{H}allucination-\textbf{A}ware \textbf{R}ecovery), a test-time recovery framework following a verify-then-steer paradigm. The verifier examines residual-stream representations of the Vision-Language Model (VLM) to identify mismatch of cross-modal evidence along the trajectories, and uses split conformal calibration to control false alarms. When an alarm is raised, the steering module uses a contextual bandit to adaptively select the rewind extent of instruction progress for online recovery. Through extensive experiments on R2R-CE and RxR-CE, our framework enhances the reliability of multiple state-of-the-art vision-language navigation methods at test time.
\end{abstract}    
\section{Introduction}

Vision-and-Language Navigation (VLN) requires an embodied agent to follow natural-language instructions and reach a target location in an unseen environment based on visual observations~\cite{r2r,rxr,VLN-CE}. Recent approaches use pretrained vision-language models (VLMs) to predict navigation actions from the instruction, the current observation, and the interaction history~\cite{zhu2026navida,cheng2024navila,zhang2024uni-navid,Liu2025navid-4D}. Yet long-horizon navigation remains vulnerable to action errors. A single wrong action can take the agent off the intended route, even leading to task failure. \textit{Reliable navigation agents need to recover from route deviations.} Most existing methods collect additional off-route trajectories and use imitation or reinforcement learning to teach navigation policies how to recover during training~\cite{an2024etpnav, activevln, dual-anchoring}. Others guide recovery with explicit reasoning capability~\cite{AwareVLN, Adanav}. Despite these advances, route deviations can still lead to navigation failures in unseen environments.

\begin{figure*}[t]
    \centering
    \includegraphics[width=0.8\textwidth]{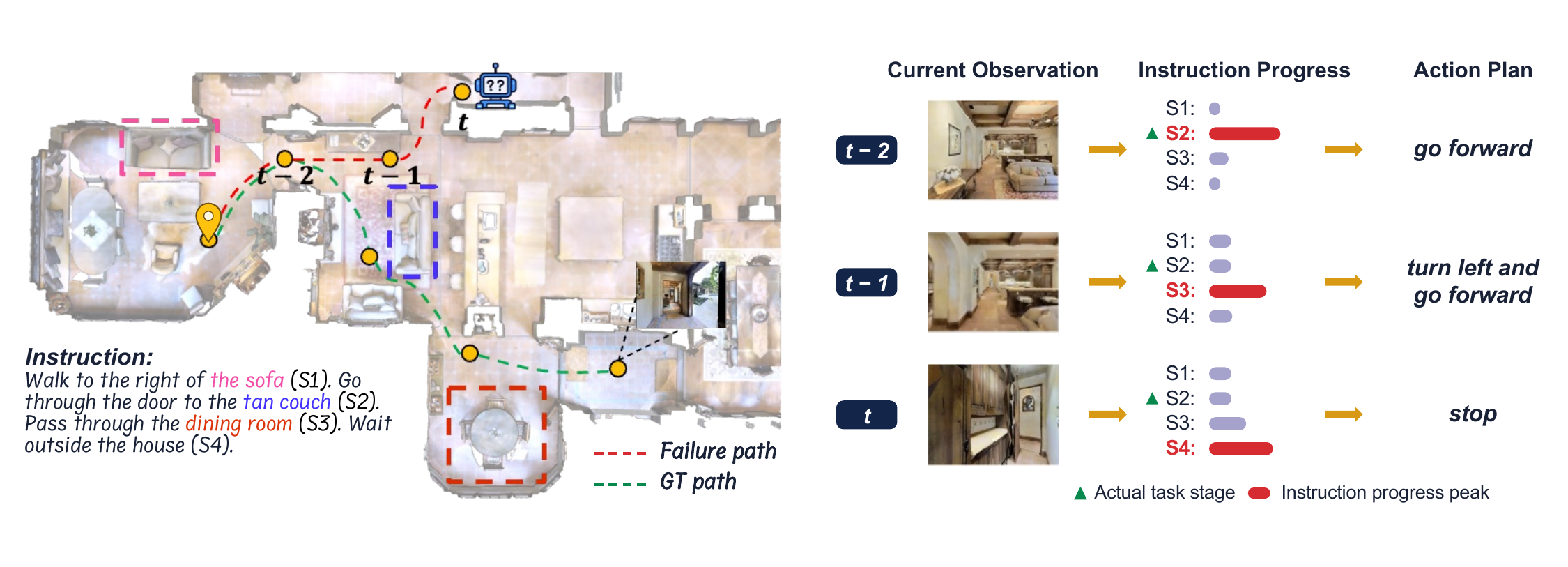}
    \caption{\textbf{An illustration of progress hallucination in VLN.}
    Ambiguity of the landmark description \textit{``tan couch''} in $S_2$ leads the agent to deviate from the reference route between $t-2$ and $t-1$.
    The resulting off-route observations are misinterpreted as evidence of task progress, causing the agent's \textcolor{red!80!black}{instruction progress} to advance from $S_2$ through $S_3$ to $S_4$, ahead of the \textcolor{green!55!black}{actual task stage}.
    At time $t$, the agent prematurely terminates based on $S_4$ (\textit{``Wait outside the house''}) despite still observing an indoor scene.}
    \label{fig:progress_hallucination}
    \vspace{-6mm}
\end{figure*}

In this work, we investigate why agents fail to recover from route deviations by examining their instruction progress, which is the attention distribution over sub-instructions at the prefill action token of the VLN agent. This distribution reveals which sub-instruction the agent relies on when taking the next move during the navigation. Through an empirical analysis of 100 failed trajectories generated by NaVIDA~\cite{zhu2026navida}, we find that VLN agents tend to misinterpret the off-route observations as evidence of task progress. As a result, their instruction progress advances beyond the actual task stage, preventing them from making recovery decisions based on the appropriate earlier sub-instructions. As illustrated in Fig.~\ref{fig:progress_hallucination}, ambiguity in the landmark description \textit{``tan couch''} causes the agent to deviate from the intended route. It then mistakes subsequently observed, visually similar indoor scenes for evidence of completing the sub-instruction \textit{``go through the dining room''}, causing its instruction progress to advance beyond the actual task stage. Consequently, the agent prematurely stops based on \textit{``Wait outside the house''} while still indoors. We term this recovery failure \textbf{progress hallucination}: The agent treats off-route observations as visual evidence of task progress, causing its instruction progress to diverge from the actual task stage and preventing it from making recovery decisions based on earlier sub-instruction. Progress hallucination substantially reduces the reliability of VLN agents in unseen environments. This raises two research questions: \textit{How can we detect navigation failure, and how can we recover from progress hallucination at test time?}

To answer the questions above,  we propose PHAR (\textbf{P}rogress \textbf{H}allucination-\textbf{A}ware \textbf{R}ecovery), a test-time recovery framework that detects navigation failure and steers instruction progress to recover from progress hallucination. For failure detection, we observe that progress hallucination can cause a mismatch between instruction progress and the visual observations. As shown in Fig.~\ref{fig:progress_hallucination}, the agent proposes to stop based on the instruction \textit{``Wait outside the house''} even though its current visual observation shows that it is still indoors. Motivated by this observation, we design a temporal probe that examines residual-stream representations at the prefill action token before termination. The probe detects navigation failure by assessing the consistency of the cross-modal evidence accumulated along the trajectory. We use split conformal prediction~\cite{CP} to calibrate the detection threshold and control false alarms. For recovery, when the probe detects navigation failure, we steer instruction progress toward an earlier stage to help the agent recover from progress hallucination. To enable the model to initiate recovery based on the appropriate earlier sub-instructions, we predefine steering actions that rewind instruction progress to varying degrees and formulate action selection as a contextual bandit problem. We use the Linear Upper Confidence Bound (LinUCB)~\cite{LinUCB} to refine steering action selection online using verifier feedback, aiming to minimize cumulative recovery regret at test time.

Overall, our contributions are threefold:
\begin{itemize}
    \item We identify progress hallucination, a failure mode in which the agent's internal instruction progress advances beyond the actual task stage, hindering recovery from route deviations.
    \item We propose PHAR, a test-time self-recovery framework following a verify-then-steer paradigm. The verifier probes residual-stream representations to identify navigation failure, while the adaptive progress steering module rewinds instruction progress to an appropriate earlier stage to support online recovery.
    \item We integrate PHAR with multiple state-of-the-art VLN methods and evaluate them on R2R-CE and RxR-CE, demonstrating consistent improvements in navigation reliability at test time.
\end{itemize}

\section{Related Work}
\subsection*{Vision-and-Language Navigation}
Vision-and-Language Navigation in Continuous Environments (VLN-CE)~\cite{Hong_2021_CVPR,VLN-CE} requires an agent to follow natural-language instructions to reach a target using egocentric observations. Recent methods leverage vision-language models (VLMs) for end-to-end navigation action prediction~\cite{zhang2024navid,cheng2024navila,activevln,zhu2026navida}. To improve navigation reliability, these methods take different approaches to correcting route deviations during training. Some expose the policy to off-route states and use imitation learning to teach it to return to the instructed route~\cite{zhu2026navida,DecoVLN}. Others sample trajectories during reinforcement learning to promote recovery from deviations~\cite{activevln}. Another approach trains the model to acquire self-aware reasoning capabilities that guide self-recovery~\cite{AwareVLN}. Despite these advances, many off-route trajectories still end in failure at test time. We investigate this limitation through instruction progress and identify progress hallucination as a failure mode that hinders recovery. Our framework PHAR addresses it at test time with a verify-then-steer paradigm.

\subsection*{Progress Estimation and Grounding in VLN}
Progress estimation and grounding help navigation agents relate instructions to their observations and execution history. Most existing methods strengthen progress tracking during training through auxiliary supervision or progress-aware reasoning. The Self-Monitoring Agent~\cite{ma2019selfmonitoring} combines visual-textual co-grounding with auxiliary progress estimation, while the Regretful Agent~\cite{ma2019regretful} uses progress estimates to guide backtracking. RCM~\cite{Wang_2019_CVPR} encourages instruction--trajectory alignment through a cross-modal matching critic. More recently, AwareVLN~\cite{AwareVLN} trains agents to explicitly assess task progress through self-aware reasoning, while SeekVLN~\cite{wang2026seek} learns to acquire supplementary visual evidence when available observations are insufficient for progress grounding. Progress supervision and grounding-oriented training aim to align instruction progress with the actual task stage, but this alignment can still break down at test time and lead to failure. Our work targets recovery after such drift has occurred.


\section{Progress Hallucination in VLN Agents}
\label{sec:progress_analysis}

In this section, we first introduce the instruction progress of the VLN agents. Then, we characterize progress hallucination in VLN agents by examining how their instruction progress diverges from the actual task stage at test time.

\subsection{Instruction Progress}
\begin{figure}[t]
    \centering
    \includegraphics[width=\columnwidth]{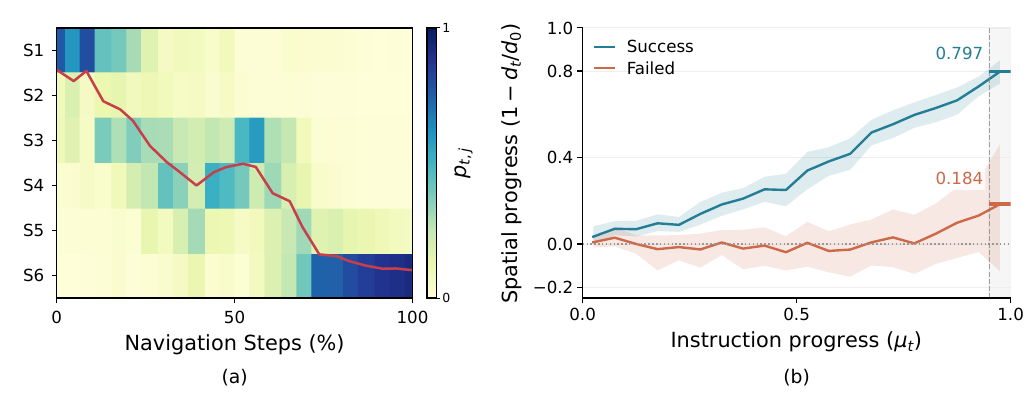}
    \caption{\textbf{Instruction progress and progress hallucination.}
    (a) Progress distribution $p_t$ (heatmap) and pointer $\mu_t$ (red curve) along a successful trajectory.
    (b) Failed trajectories achieve less spatial progress than successful trajectories at comparable $\mu_t$. Spatial progress is $1-d_t/d_0$, where $d_t$ and $d_0$ are the current and initial goal distances.}
    \label{fig:progress_analysis_ab}
\end{figure}

At decision step $t$, a VLN agent's VLM predicts actions from a navigation instruction $I$, the current visual observation $o_t$, and the sequence of historical observations $\mathcal H_t$. We define the pre-action position as the final prefill
position immediately preceding the first generated action token. We refer to the residual-stream representation at this position as the VLM state $\bm v_t$. We characterize instruction progress using attention from $\bm v_t$ to the sub-instruction representations, reflecting which part of the instruction the agent attends to when selecting its next action.
Given $K>1$ ordered sub-instructions $(S_1,\ldots,S_K)$, we aggregate and normalize the VLM's attention to obtain the instruction progress distribution $p_t=(p_{t,1},\ldots,p_{t,K})$:
\begin{equation}
    p_{t,j} = \frac{\alpha_t(S_j)}
    {\sum_{\ell=1}^{K}\alpha_t(S_\ell)},
    \label{eq:progress_distribution}
\end{equation}
where $\alpha_t(S_j)$ denotes the head-averaged attention mass assigned to sub-instruction $S_j$ at a fixed decoder layer, and $p_{t,j}$ represents the normalized attention allocated to $S_j$ among all sub-instructions at decision step $t$.
We further define the progress pointer $\mu_t\in[0,1]$ as the distribution's center of mass over normalized sub-instruction positions:
\begin{equation}
    \mu_t = \sum\nolimits_{j=1}^{K} p_{t,j}\frac{j-1}{K-1}.
    \label{eq:progress_pointer}
\end{equation}
The distribution $p_t$ describes how attention is allocated across sub-instructions, while $\mu_t$ summarizes the agent's focus along the instruction, with larger values indicating later stages.
Fig~\ref{fig:progress_analysis_ab}~(a) illustrates the normal transition of $\mu_t$ and $p_t$ along a successful trajectory: each heatmap column shows $p_t$, and the red curve traces $\mu_t$ in sub-instruction coordinates.
As navigation proceeds, attention $p_t$ shifts from the first sub-instruction $S_1$ toward the final one $S_6$, moving the pointer $\mu_t$ toward the end of the instruction.

\subsection{Investigation of Progress Hallucination}
To investigate progress hallucination, we conduct three experiments on 100 failed NaVIDA~\cite{zhu2026navida} trajectories spanning 44 tasks and 10 scenes from the val-unseen split of R2R-CE~\cite{r2r,VLN-CE}. Throughout the analysis, we extract $p_t$ and progress pointer $\mu_t$ from the decoder layer $-12$ of NaVIDA. Our first experiment characterizes progress hallucination with $\mu_t$. The second investigates why it occurs. The third experiment analyzes how it can be mitigated.
\paragraph{Characterizing Progress Hallucination.}
To characterize progress hallucination, we first analyze the relationship between instruction progress and spatial progress along successful reference-following and failed trajectories.
Instruction progress is measured by the pointer $\mu_t$, while spatial progress is assessed using the agent's normalized distance to the goal. As shown in Fig.~\ref{fig:progress_analysis_ab}~(b), successful trajectories approach the goal as the pointer advances, whereas failed trajectories make little progress toward the goal despite advancing instruction progress.
For $\mu_t\in[0.95,1]$, mean spatial progress is only 0.184 for failed trajectories, compared with 0.797 for successful trajectories. This mismatch reveals progress hallucination: the agent attends to later sub-instructions without making corresponding progress toward the goal.

\begin{figure}[t]
    \centering
    \includegraphics[width=\columnwidth]{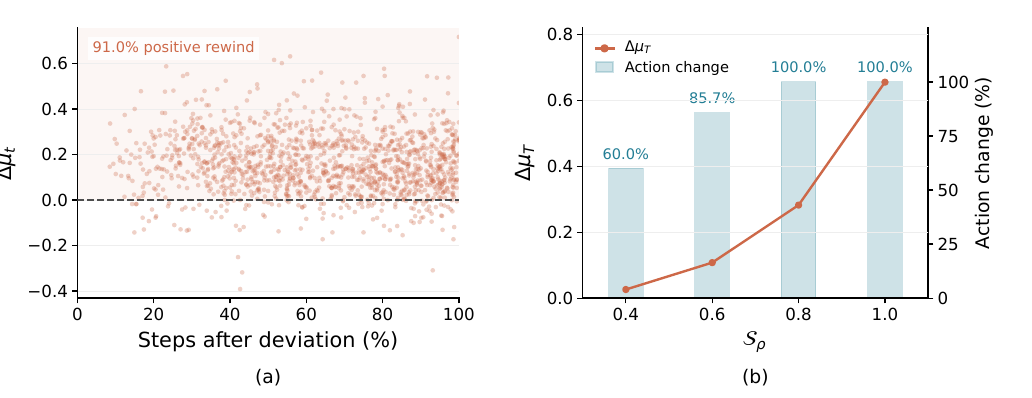}
    \caption{\textbf{Investigating the cause and mitigation of progress hallucination.}
    (a) Each point shows pointer rewind $\Delta\mu_t$ under gray-image replacement of off-route history. Steps are normalized from deviation onset to termination.
    (b) Early-history truncation induces pointer rewind $\Delta\mu_T$ (left axis) and changes predicted actions from \texttt{STOP} to movement (right axis).}
    \label{fig:progress_analysis_cd}
\end{figure}

\paragraph{Why does Progress Hallucination Occur?}
We hypothesize that progress hallucination arises because the agent treats observations acquired off-route as evidence of task progress.
To test this explanation, we measure the sensitivity of instruction progress to these off-route observations in failed trajectories.
We identify deviation onset as the first of three consecutive navigation steps more than one meter from the reference path.
After deviation onset, we replace sampled history images from the preceding ten decisions with uniform gray images, considering only interventions where all replaced images were captured off-route.
We compute the pointer difference before and after replacement as $\Delta\mu_t=\mu_t-\widetilde\mu_t$, where $\widetilde\mu_t$ is the pointer obtained with the intervened history.
A positive value indicates that the replaced off-route observations supported the advancement of instruction progress.
As shown in Fig.~\ref{fig:progress_analysis_cd}(a), over 91\% of the evaluated interventions yield positive rewind across the post-deviation phase.
This suggests that the agent continues to interpret off-route observations as evidence of task progress, giving rise to progress hallucination.

\paragraph{How to Mitigate Progress Hallucination?}
Building on the cause of progress hallucination, we next explore how to steer the progress pointer back to early stages to mitigate progress hallucination. Inspired by the biological phenomenon of forgetting, which can reduce the influence of outdated information on decision-making~\cite{Forget}, we explore whether truncating early visual history can mitigate the premature advancement of instruction progress driven by off-route observations. Specifically, we intervene at the termination step $T$ of each failed trajectory. We define a set of progress steering operators $\mathcal S=\{\mathcal S_\rho\mid\rho\in\{0.4,0.6,0.8,1.0\}\}$, where each $\mathcal S_\rho$ removes the earliest fraction $\rho$ of observations from $\mathcal H_T$.
At this fixed state, we apply each operator and measure the pointer rewind extent as $\Delta\mu_T=\mu_T-\widetilde\mu_T$, where $\widetilde\mu_T$ is the pointer obtained with the truncated history $\mathcal S_\rho(\mathcal H_T)$.
Fig~\ref{fig:progress_analysis_cd}~(b) shows that mean rewind increases from 0.026 under 40\% deletion to 0.655 under full deletion, accompanied by a rise in the action change rate (the fraction of cases switching from \texttt{STOP} to a movement action) from 60.0\% to 100\%. These results suggest that the progress steering operators $\mathcal S_\rho$ rewind the progress pointer, mitigating progress hallucination. This steering also changes the model's predicted actions, suggesting that the operator can prompt the agent to take actions that creates opportunities for recovery.

\section{Method}
\label{sec:method}

\begin{figure*}[t]
    \centering
    \includegraphics[width=0.9\textwidth]{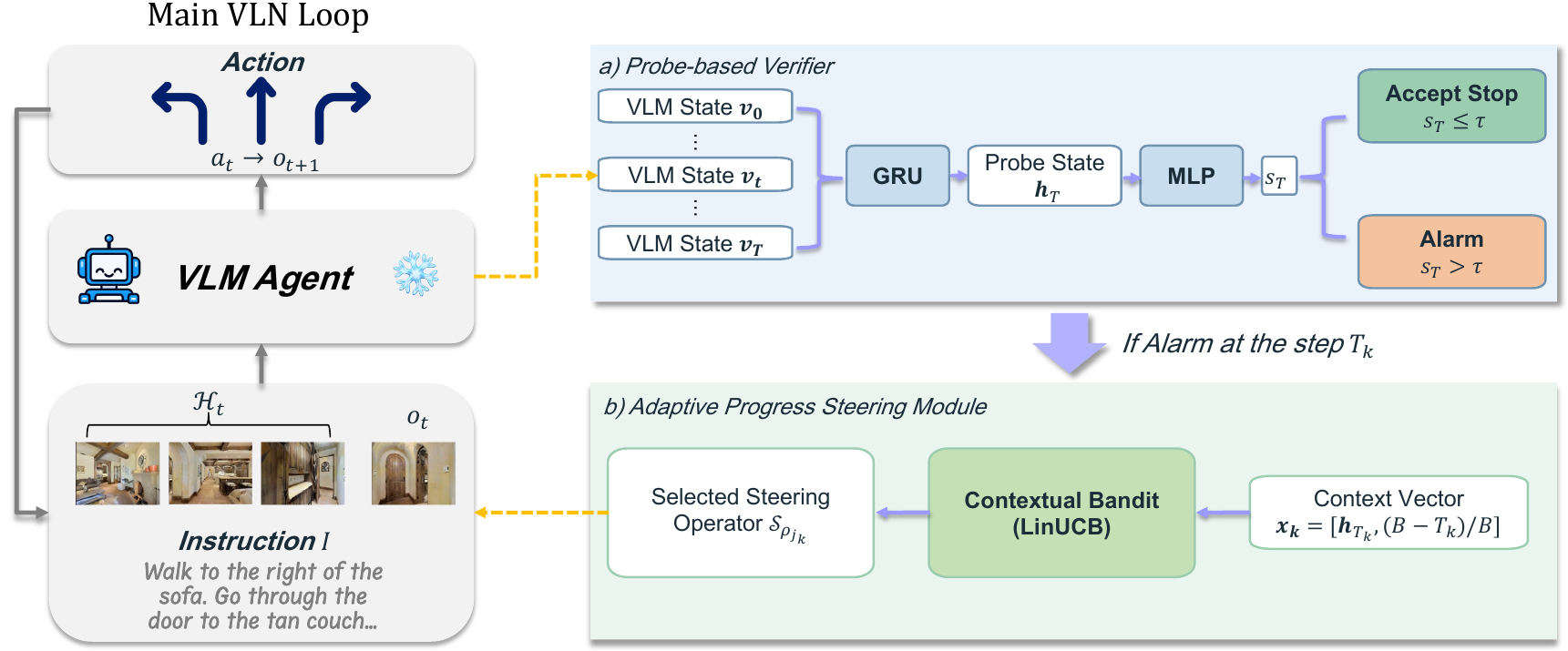}
    \caption{\textbf{Overview of our PHAR framework.}
    \textit{\textbf{(a) Probe-Based Verifier:} a temporal probe aggregates VLM states} $\bm v_t$ \textit{into a probe state} $\bm h_t$\textit{ to score a proposed} \texttt{STOP}\textit{; the calibrated threshold} $\tau$ \textit{determines whether to raise an alarm. \textbf{(b) Adaptive Progress Steering Module:} upon an alarm at the step} $T_k$\textit{, LinUCB selects an appropriate progress steering operator }$\mathcal S_{\rho_{j_k}}$ \textit{using the probe state} $\bm h_{T_k}$ \textit{and remaining step budget.}}
    \label{fig:method_framework}
\end{figure*}

To address navigation failure caused by progress hallucination, we propose a test-time recovery framework PHAR that operates outside the navigation agent comprising two modules. As shown in Fig.~\ref{fig:method_framework}, when the agent proposes a \texttt{STOP} action at the decision step $T$, the \emph{Probe-Based Verifier} detects navigation failure by assessing the consistency of cross-modal evidence accumulated along the trajectory.
If the verifier raises an alarm, we intercept the proposed \texttt{STOP} action and invoke the \emph{Adaptive Progress Steering Module} to steer the instruction progress pointer $\mu_T$ by selecting an appropriate steering operator $\mathcal S_\rho$ for online recovery.

\subsection{Problem Formulation}

We consider vision-and-language navigation in unseen continuous 3D environments~\cite{VLN-CE}.
The agent is given a natural-language instruction $ I=(w_1,\ldots,w_L)$ describing a route from its starting position to a target location. 
At each navigation decision $t$, it receives an egocentric RGB observation $ o_t$ and maintains a visual history $\mathcal H_t$ of preceding observations.
Its goal is to follow the instruction and stop at the specified destination by executing low-level actions from a discrete action set $\mathcal A=\{\texttt{FORWARD},\texttt{TURN-LEFT},\texttt{TURN-RIGHT},\texttt{STOP}\}$.
To select these actions, a pretrained VLM takes $(I, o_t, \mathcal H_t)$ as input and generates a textual action plan describing the next movement (e.g, "turn right 15 degrees"), which is parsed into one or more executable low-level actions $a_{t+1:t+k}$. The VLM agent executes these actions, obtains new observations, and repeats this process to navigate effectively in unseen environments, under the decision step budget $B$.

\subsection{Probe-Based Verifier}
\label{sec:verifier}

As illustrated in Fig.~\ref{fig:progress_hallucination}, progress hallucination can cause an agent to terminate despite
a mismatch between its visual observations and instruction progress. Detecting this mismatch is key to identifying navigation failure. Recent studies suggest that internal-state probes can identify visually unsupported VLM outputs~\cite{halp-probe, UniProbe}. Inspired by these studies, we design the Probe-Based Verifier to assess termination reliability from the sequence of VLM states $\bm v_t$, which can encode cross-modal inconsistencies accumulated during navigation.

\paragraph{Temporal Probe $\psi$.}
Specifically, the temporal probe projects each VLM state $\bm v_t$ through a linear projection $W_p$ and uses a causal GRU to aggregate the projected states across decisions into a probe state $\bm h_t$:
\begin{equation}
    \bm{h}_t=\operatorname{GRU}_\psi(W_p \bm {v}_t,\bm{h}_{t-1}),
    \label{eq:verifier_state}
\end{equation}
and an MLP $F_\psi$ produces the failure score with a sigmoid activation function:
\begin{equation}
    s_t=\operatorname{sigmoid}\!\left(F_\psi([W_p \bm {v}_t;\bm{h}_t])\right).
    \label{eq:verifier_score}
\end{equation}

We train the temporal probe using trajectory outcomes, with $y=1$ for failure and $y=0$ for success.
For each trajectory, the probe processes the sequence of residual-stream representations and receives supervision at the time step $T$, when the agent proposes the \texttt{STOP} decision.
We minimize the binary cross-entropy loss over $N_v$ training trajectories:
\begin{equation}
    \mathcal L_{\mathrm{ver}}
    = -\frac{1}{N_v}
    \sum_{i=1}^{N_v}
    \left[
        y_i \log s_T^i
        + (1-y_i)\log\!\left(1-s_T^i\right)
    \right].
    \label{eq:verifier_loss}
\end{equation}

\paragraph{Conformal Calibration.}
To limit unnecessary recovery on successful trajectories, we use split conformal calibration to determine the alarm threshold $\tau$ from held-out successful trajectories~\cite{CP}. 
Let $\mathcal C$ contain these scores and let $n_{\mathrm{cal}}=|\mathcal C|$.
For a target false-alarm level $\alpha_{\mathrm{cal}}$, we compute the quantile level
\begin{equation}
    q_{\mathrm{cal}}=\min\!\left\{1,
    \frac{\lceil(n_{\mathrm{cal}}+1)(1-\alpha_{\mathrm{cal}})\rceil}
         {n_{\mathrm{cal}}}\right\}.
    \label{eq:verifier_calibration}
\end{equation}
We set $\tau=Q_{q_{\mathrm{cal}}}^{\uparrow}(\mathcal C)$,
where $Q_{q_{\mathrm{cal}}}^{\uparrow}$ denotes the upper empirical quantile at level $q_{\mathrm{cal}}$.
When the agent proposes \texttt{STOP} at time step $T$, we accept it if $s_T\le\tau$.
If $s_T>\tau$ and the execution budget $B$ has not been exhausted, the following Adaptive Progress Steering Module selects an appropriate progress steering operator to recover from hallucination.

\subsection{Adaptive Progress Steering Module}
\label{sec:steering}

When the verifier rejects termination at step $T$ ($s_T>\tau$), we apply $\mathcal S_\rho$ to remove the earliest fraction $\rho$ of history observations $\mathcal H_T$ (Sec.~\ref{sec:progress_analysis}). This rewinds $\mu_T$ to encourage action reasoning based on appropriate earlier sub-instructions. Different degrees of premature instruction progress require different steering strengths: Excessive steering may cause the agent to follow earlier sub-instructions and explore unnecessarily, whereas insufficient steering may leave progress hallucination unresolved and hinder recovery.
We therefore propose the Adaptive Progress Steering Module to adaptively select a progress steering operator from $\mathcal S=\{\mathcal S_{\rho_j}\}_{j=1}^{M}$ with different steering strengths $\rho_j$ for different failure cases.

\subsubsection{Formulation of Progress Steering}
After progress steering, the agent may execute multiple navigation steps before recovery feedback becomes available.
We formulate steering operator selection as a reinforcement learning problem in the contextual bandit setting~\cite{LinUCB}: the selector chooses a steering action based on the current context and learns from the resulting recovery reward, treating the intervening navigation as a black-box process.
We define its context vectors, steering action space, and optimization objective below.

\paragraph{Contextual Information.}
To enable the bandit to select an appropriate steering operator, its context should encode the reliability of the current trajectory and the decision steps available for subsequent recovery.
Specifically, for the $k$-th trajectory entering recovery, let $T_k$ denote the decision step that the proposed \texttt{STOP} triggers the verifier's alarm. The context of the bandit is $\bm{x}_k=[\bm{h}_{T_k};(B-T_k)/B]$, where $\bm{h}_{T_k}$ is the temporal probe state of the trajectory and $B$ is the per-episode decision step budget. This vector $\bm{x}_k$ summarizes the execution state and remaining budget to guide steering action selection for recovery.

\paragraph{Steering Action Space.}
We define the bandit's discrete steering action space as $\mathcal U=\{\bm{u}_j\}_{j=1}^{M}$, where each one-hot vector $\bm{u}_j$ encodes a progress steering operation $\mathcal S_{\rho_j}$.
Selecting $\bm{u}_j$ applies $\mathcal S_{\rho_j}(\mathcal H_{T_k})$, removing the earliest fraction $\rho_j$ of history while retaining the current observation. As shown in Fig.~\ref{fig:progress_analysis_cd}~(b), the truncation ratio $\rho_j$ controls steering strength: removing more early history produces greater pointer rewind.

\paragraph{Optimization Objective.}
The bandit selects a steering action $\bm{u}_k\in\mathcal U$ based on the context vector $\bm{x}_k$ and refines subsequent selections using the online reward $r_k$.
This reward is characterized by a joint function of recovery quality and decision step cost, as defined in Eq.~\eqref{eq:recovery_reward} in the next subsection.
Our objective is to minimize cumulative recovery regret across test trajectories under the per-episode decision step budget $B$:
\begin{equation}
    R_K=\mathbb E\!\left[\sum_{k=1}^{K}
    \bigl(r(\bm{x}_k,\bm{u}_k^*)-r(\bm{x}_k,\bm{u}_k)\bigr)\right],
    \label{eq:recovery_regret}
\end{equation}
where $r(\bm x,\bm u)$ denotes the expected recovery reward for context $\bm x$ and action $\bm u$, and $\bm u_k=\bm u_{j_k}$ is the selected steering action. Here, $\bm u_k^*$ denotes the optimal steering action for context $\bm x_k$. By minimizing this regret, the bandit learns to select steering actions with greater expected recovery benefit for different alarmed trajectories.

\begin{algorithm}[t]
\caption{Adaptive Progress Steering via LinUCB}
\label{alg:self_recovery}
\small
\begin{algorithmic}[1]
\Require Test episodes; steering actions $\mathcal U$; alarm threshold $\tau$; step budget $B$
\State $k\gets1$; $\bm{A}_1\gets\lambda\bm{I}_d$; $\bm{b}_1\gets\bm{0}$
\For{each test episode}
    \State Reset navigation history and probe state
    \State Navigate until a proposed $\texttt{STOP}$ action at step $T_k$
    \If{$s_{T_k}>\tau$ and $T_k<B$}
        \State $\bm{x}_k\gets[\bm{h}_{T_k};(B-T_k)/B]$
        \State $\bm{x}_{k,j}\gets\bm{u}_j\otimes\bm{x}_k$ for each $\bm{u}_j\in\mathcal U$
        \State $\widehat{\bm{w}}_k\gets\bm{A}_k^{-1}\bm{b}_k$
        \Statex \Comment{\textcolor{gray}{Select a steering action via LinUCB}}
        \State $\displaystyle j_k\gets\arg\max_j\left[\bm{x}_{k,j}^{\top}\widehat{\bm{w}}_k+\beta\sqrt{\bm{x}_{k,j}^{\top}\bm{A}_k^{-1}\bm{x}_{k,j}}\right]$
        \State Apply Steering Operator $\mathcal S_{\rho_{j_k}}$ ($\mathcal H_{T_k}\gets \mathcal S_{\rho_{j_k}}(\mathcal H_{T_k})$)
        \State Resume navigation to the next proposed \texttt{STOP} at $T_k'$
        \State Compute $r_k$ using Eq.~\eqref{eq:recovery_reward}
        \Statex \Comment{\textcolor{gray}{Update LinUCB with recovery feedback}}
        \State $\bm{A}_{k+1}\gets\bm{A}_k+\bm{x}_{k,j_k}\bm{x}_{k,j_k}^{\top}$
        \State $\bm{b}_{k+1}\gets\bm{b}_k+r_k\bm{x}_{k,j_k}$
        \State $k\gets k+1$
    \Else
        \State Terminate the episode
    \EndIf
\EndFor
\end{algorithmic}
\end{algorithm}

\subsubsection{Adaptive Progress Steering via Bandit}
Building on the formulation above, we implement online adaptive progress steering using Linear Upper Confidence Bound (LinUCB)~\cite{LinUCB}, as summarized in Algorithm~\ref{alg:self_recovery}.
LinUCB balances exploitation and exploration by selecting steering actions according to their upper confidence scores.

Specifically, for the $k$-th trajectory entering recovery, we encode each context--action pair as $\bm{x}_{k,j}=\bm{u}_j\otimes\bm{x}_k\in\mathbb R^d$, where $\otimes$ denotes the Kronecker product.
LinUCB estimates expected reward using $\widehat{\bm{w}}_k=\bm{A}_k^{-1}\bm{b}_k$ and selects the action $\bm{u}_{j_k}$ with the highest upper confidence score.
We apply its corresponding operator $\mathcal S_{\rho_{j_k}}$ at decision step $T_k$ and resume navigation until the next proposed \texttt{STOP} at $T_k'$ within the same episode.
The verifier then provides feedback on the resulting termination.
If its alarm clears, the action receives a reliability reward of $1-s_{T_k'}$; otherwise, it receives a penalty of $-\eta$, penalizing recovery steps:
\begin{equation}
    r_k=
    \begin{cases}
        1-s_{T_k'}-\gamma\dfrac{T_k'-T_k}{B}, & s_{T_k'}\le\tau,\\
        -\eta-\gamma\dfrac{T_k'-T_k}{B}, & s_{T_k'}>\tau,
    \end{cases}
    \label{eq:recovery_reward}
\end{equation}
where $\eta>0$ discourages ineffective steering and $\gamma>0$ balances termination reliability against decision step cost. After observing $r_k$, we update the feature matrix $\bm{A}_k$ and feature--reward vector $\bm{b}_k$ using the selected context--action feature.
The parameters $\lambda>0$ and $\beta>0$ control regularization and exploration, respectively.
Bandit statistics are retained across test trajectories, allowing subsequent steering action selections to benefit from accumulated recovery feedback online.

\section{Experiments}

\label{sec:experiments}

We evaluate PHAR on multiple state-of-the-art navigation baselines. We present the experimental setup and main results, followed by component ablations and qualitative trajectory visualizations.

\subsection{Experimental Setup}
\label{sec:experimental_setup}

\begin{table*}[t]
\centering
\caption{Comparison with existing methods on the R2R-CE~\cite{VLN-CE} and RxR-CE~\cite{rxr}
benchmarks. $*$ denotes methods based on the waypoint predictor from~\cite{Waypoint_predictor}; $\dagger$ denotes our reproduction results. Shaded rows show our results for NaVILA~\cite{cheng2024navila}, NaVIDA~\cite{zhu2026navida}, and AwareVLN~\cite{AwareVLN}. The best and second-best results in each metric column
are \textbf{bolded} and \underline{underlined}, respectively.}
\label{tab:main_results}

\begingroup
\setlength{\tabcolsep}{5pt}
\renewcommand{\arraystretch}{0.83}
\fontsize{5.3}{7.7}\selectfont
\resizebox{0.9\textwidth}{!}{%
\begin{tabular}{@{}lccccccccccc@{}}
\toprule
\multirow{2}{*}{Method}
& \multicolumn{4}{c}{Observation}
& \multicolumn{4}{c}{R2R-CE Val-Unseen}
& \multicolumn{3}{c}{RxR-CE Val-Unseen} \\
\cmidrule(lr){2-5}
\cmidrule(lr){6-9}
\cmidrule(lr){10-12}
& S.RGB & Pano. & Depth & Odo.
& NE$\downarrow$ & OSR$\uparrow$ & SR$\uparrow$ & SPL$\uparrow$
& NE$\downarrow$ & SR$\uparrow$ & SPL$\uparrow$ \\
\midrule

Sim2Sim$^{*}$~\cite{sim-to-sim}
& & \checkmark & \checkmark & \checkmark
& 6.07 & 52.0 & 43.0 & 36.0 & 8.76 & 26.5 & 22.1 \\
GridMM$^{*}$~\cite{wang2023gridmm}
& & \checkmark & \checkmark & \checkmark
& 5.11 & 61.0 & 49.0 & 41.0 & -- & -- & -- \\
DreamWalker$^{*}$~\cite{Dreamwalker}
& & \checkmark & \checkmark & \checkmark
& 5.53 & 59.0 & 49.0 & 44.0 & -- & -- & -- \\
Reborn$^{*}$~\cite{Reborn}
& & \checkmark & \checkmark & \checkmark
& 5.40 & 57.0 & 50.0 & 46.0 & 5.98 & 48.6 & 42.0 \\
ETPNav$^{*}$~\cite{an2024etpnav}
& & \checkmark & \checkmark & \checkmark
& 4.71 & 65.0 & 57.0 & 49.0 & 5.64 & 54.7 & 44.8 \\
HNR$^{*}$~\cite{Wang_2024_CVPR_HNR}
& & \checkmark & \checkmark & \checkmark
& 4.42 & 67.0 & 61.0 & 51.0 & 5.50 & 56.3 & 46.7 \\
\midrule

Navid~\cite{zhang2024navid}
& \checkmark & & &
& 5.47 & 49.0 & 37.0 & 35.0 & 8.41 & 23.8 & 21.2 \\
Uni-Navid~\cite{zhang2024uni-navid}
& \checkmark & & &
& 5.58 & 53.5 & 47.0 & 42.7 & 6.24 & 48.7 & 40.9 \\
StreamVLN~\cite{wei2025streamvln}
& \checkmark & & &
& 4.98 & 64.2 & 56.9 & 51.9 & 6.22 & 52.9 & 46.0 \\
ActiveVLN~\cite{activevln}
& \checkmark & & &
& 5.31 & 58.3 & 50.1 & 43.7 & 5.84 & 50.7 & 41.2 \\
MonoDream~\cite{wang2025monodream}
& \checkmark & & &
& 5.45 & 61.5 & 55.8 & 49.1 & 6.38 & 49.4 & 40.9 \\
NavFoM~\cite{NavFoM}
& \checkmark & & &
& 5.01 & 64.9 & 56.2 & 51.2 & 5.51 & 57.4 & 49.4 \\
Progress-Think~\cite{Wang_2026_CVPR}
& \checkmark & & &
& 4.68 & 63.6 & 60.1 & 53.6 & -- & -- & -- \\
AdaNav-NaVILA~\cite{Adanav}
& \checkmark & & &
& 5.01 & 66.6 & 60.2 & 50.0 & 6.21 & 60.5 & 49.8 \\
\midrule

NaVILA~\cite{cheng2024navila}
& \checkmark & & &
& 5.22 & 62.5 & 54.0 & 49.0 & 6.77 & 49.3 & 44.0 \\
\rowcolor{oursblue}
\textbf{NaVILA+PHAR}\textsubscript{ours}
& \checkmark & & &
& 4.78
& 71.0
& 60.7
& 52.1
& 5.55
& 59.3
& 48.4 \\
\midrule

NaVIDA~\cite{zhu2026navida}
& \checkmark & & &
& 4.32 & 69.5 & 61.4 & 54.7 & 5.23 & 57.4 & 49.6 \\
\rowcolor{oursblue}
\textbf{NaVIDA+PHAR}\textsubscript{ours}
& \checkmark & & &
& \underline{3.98}
& \underline{78.6}
& \underline{67.0}
& 54.8
& 4.96
& 63.2
& 50.4\\
\midrule

AwareVLN~\cite{AwareVLN}
& \checkmark & & &
& 4.02 & 73.5 & 65.4 & \underline{55.1}
& \underline{3.95} & \underline{67.6} & \textbf{56.1} \\
AwareVLN$^{\dagger}$~\cite{AwareVLN}
& \checkmark & & &
& 4.02 & 74.5 & 65.2 & 55.0 & 4.21 & 65.7 & 54.4 \\
\rowcolor{oursblue}
\textbf{AwareVLN+PHAR}\textsubscript{ours}
& \checkmark & & &
& \textbf{3.67}
& \textbf{79.3}
& \textbf{68.5}
& \textbf{57.4}
& \textbf{3.89}
& \textbf{68.7}
& \underline{54.9}\\
\bottomrule
\end{tabular}%
}

\endgroup
\end{table*}

\paragraph{Benchmarks.}
R2R-CE and RxR-CE~\cite{VLN-CE} extend the original R2R and RxR datasets~\cite{r2r,rxr} to continuous navigation by converting discrete reference trajectories into executable paths in the Habitat simulator~\cite{habitat}.
Both benchmarks use indoor 3D environments from Matterport3D~\cite{Matterport3D}.
Their val-unseen splits on R2R-CE and RxR-CE contain 1,839 and 11,006 episodes, respectively.
R2R-CE provides English instructions, whereas RxR-CE includes instructions in three languages.
Compared with R2R-CE, RxR-CE features longer trajectories and more detailed instructions, making it more challenging for instruction following.

\paragraph{Implementation Details.}
We apply PHAR to multiple state-of-the-art methods~\cite{zhu2026navida,cheng2024navila,AwareVLN}, keeping their VLMs frozen.
We train the verifier offline for each method with 2,000 successful and 2,000 failed trajectories from the training split.
Using held-out successful val-seen trajectories, we perform split conformal calibration with $\alpha_{\mathrm{cal}}=0.03$ to select $\tau$, then fix the verifier and threshold for val-unseen evaluation.
We use five progress steering strengths $\rho\in\{0.4,0.6,0.8,0.9,1.0\}$, each specifying the fraction of early visual history removed by $\mathcal S_\rho$.

\paragraph{Evaluation Metrics.}
Following standard VLN evaluation~\cite{VLN-CE,VLN_metric_agent}, we use success rate (SR) and success weighted by path length (SPL) as the primary metrics.
SR measures the proportion of episodes in which the agent stops within 3\,m of the goal, while SPL additionally accounts for path efficiency.
We also report navigation error (NE), the remaining distance to the goal at termination, and oracle success rate (OSR), the proportion of trajectories that reach within 3\,m of the goal at any point.
Together, these metrics assess recovery success and navigation efficiency.

\subsection{Main Results}
\label{sec:main_results}
Table~\ref{tab:main_results} shows that PHAR consistently improves three navigation baselines on the val-unseen splits of R2R-CE and RxR-CE while keeping their VLMs frozen and retaining monocular RGB inputs. For NaVILA~\cite{cheng2024navila}, NaVIDA~\cite{zhu2026navida}, and AwareVLN~\cite{AwareVLN}, SR increases by 6.7, 5.6, and 3.3 percentage points on R2R-CE, and by 10.0, 5.8, and 3.0 points on RxR-CE, respectively. We further compare PHAR with AdaNav-NaVILA~\cite{Adanav}, which finetunes the agent through reinforcement learning on additional interaction trajectories to enable adaptive multi-mode reasoning. Without finetuning the navigation model, PHAR achieves comparable R2R-CE SR (60.7\% versus 60.2\%) and higher SPL (52.1\% versus 50.0\%), highlighting the effectiveness of test-time progress steering for navigation recovery.

\subsection{Ablation Study}
We conduct ablation studies on NaVIDA using the R2R-CE and RxR-CE val-unseen splits to evaluate the effectiveness of the probe-based verifier for failure detection at test time and the necessity of adaptive steering-strength selection.

\begin{table}[htbp]
\centering
\caption{Ablation study on probe-based verifier. Failure precision (P) and recall (R) use successful val-seen calibration at $\alpha_{\mathrm{cal}}=0.03$. All metrics are percentages.}
\label{tab:verifier_baselines}

\begingroup
\fontsize{7}{8.2}\selectfont
\setlength{\tabcolsep}{3pt}
\renewcommand{\arraystretch}{1.0}

\resizebox{0.9\columnwidth}{!}{%
\begin{tabular}{@{}lcccccc@{}}
\toprule
\multirow{2}{*}{Method}
& \multicolumn{3}{c}{R2R-CE Val-Unseen}
& \multicolumn{3}{c}{RxR-CE Val-Unseen} \\
\cmidrule(lr){2-4}
\cmidrule(lr){5-7}
& P$\uparrow$ & R$\uparrow$ & AUROC$\uparrow$
& P$\uparrow$ & R$\uparrow$ & AUROC$\uparrow$ \\
\midrule
NLL
& 43.24 & 4.99 & 52.05
& 51.96 & 4.09 & 54.83 \\
Entropy
& 31.43 & 1.72 & 51.29
& 44.55 & 3.47 & 54.00 \\
Cosine $k$NN
& 73.08 & 17.78 & 72.30
& 80.93 & 16.04 & 74.02 \\
Prefill MLP
& 82.57 & 31.05 & 78.35
& 82.98 & 24.44 & 80.11 \\
\midrule
\rowcolor{cyan!12}
\textbf{Verifier}\textsubscript{ours}
& \textbf{83.89} & \textbf{47.11} & \textbf{85.09}
& \textbf{90.06} & \textbf{45.41} & \textbf{88.16} \\
\bottomrule
\end{tabular}%
}
\endgroup
\end{table}

\paragraph{Ablation on Probe-Based Verifier.}
Table~\ref{tab:verifier_baselines} evaluates two aspects of our verifier: the value of VLM internal states for failure detection and the benefit of GRU-based temporal modeling.
\textit{Internal states.} We use \texttt{STOP} negative log-likelihood (NLL) and entropy as output-based measures of decoding confidence and uncertainty, respectively.
Both achieve near-chance AUROC on R2R-CE and RxR-CE, indicating that output confidence alone poorly distinguishes successful from failed trajectories.
These results support examining VLM internal states for mismatches between instruction progress and visual observations, providing more reliable information for failure detection.
\textit{Temporal modeling.} We compare our verifier with two baselines that detect failure using only the internal representation at the pre-action position of the termination step: cosine $k$NN for anomaly detection and an MLP for supervised failure classification. Our verifier outperforms both, improving AUROC over the MLP from 78.35\% to 85.09\% on R2R-CE and from 80.11\% to 88.16\% on RxR-CE. By aggregating cross-modal inconsistency signals across multiple decision steps, our GRU-based temporal probe provides more reliable failure detection than these single-step baselines.

\begin{table}[htbp]
\centering
\caption{Ablation study of Adaptive Progress Steering Module on R2R-CE and RxR-CE Val-Unseen splits.}
\label{tab:phar_steering_ablation}

\begingroup
\fontsize{8}{8.2}\selectfont
\setlength{\tabcolsep}{3pt}
\renewcommand{\arraystretch}{1.08}

\begin{tabular}{lcccccc}
\toprule
\multirow{2}{*}{Method}
& \multicolumn{3}{c}{R2R-CE Val-Unseen}
& \multicolumn{3}{c}{RxR-CE Val-Unseen} \\
\cmidrule(lr){2-4}
\cmidrule(lr){5-7}

& NE$\downarrow$ & SR$\uparrow$ & SPL$\uparrow$
& NE$\downarrow$ & SR$\uparrow$ & SPL$\uparrow$ \\
\midrule
 NaVIDA~\cite{zhu2026navida}
& 4.32 & 61.4 & 54.7
& 5.23 & 57.4 & 49.6 \\

w/o Steering
& 4.54 & 63.3 & 53.9
& 5.29 & 59.7 & 48.9 \\

 w/o Bandit 
& 4.07 & 64.7 & 54.0
& 5.21 & 60.3 & 49.4 \\

\midrule
\rowcolor{cyan!12}
 \textbf{PHAR}\textsubscript{ours}
& \textbf{3.98} & \textbf{67.0} & \textbf{54.8}
& \textbf{4.96} & \textbf{63.2} & \textbf{50.4} \\
\bottomrule
\end{tabular}
\endgroup
\end{table}

\paragraph{Ablation on Progress Steering Module.}
We conduct two ablations to assess the importance of progress steering and the necessity of bandit-based adaptive selection (Table~\ref{tab:phar_steering_ablation}).
To evaluate progress steering, we compare PHAR with \textit{w/o Steering}, which retains the verifier and resumes navigation with visual history unchanged.
To evaluate adaptive selection, we compare PHAR with \textit{w/o Bandit}, which uniformly samples a history-truncation strength without online optimization.
Without progress steering, SR falls from 67.0\% to 63.3\% on R2R-CE and from 63.2\% to 59.7\% on RxR-CE, showing that rejecting termination alone can not resolve hallucination, hindering navigation recovery.
Random steering improves SR over \textit{w/o Steering}, but remains below PHAR at 64.7\%/60.3\% on R2R-CE/RxR-CE.
This further supports adapting selection of steering strength through online feedback rather than just selecting randomly.

\subsection{Rollout Visualization}

\begin{figure}[t]
    \centering
    \includegraphics[width=\columnwidth]{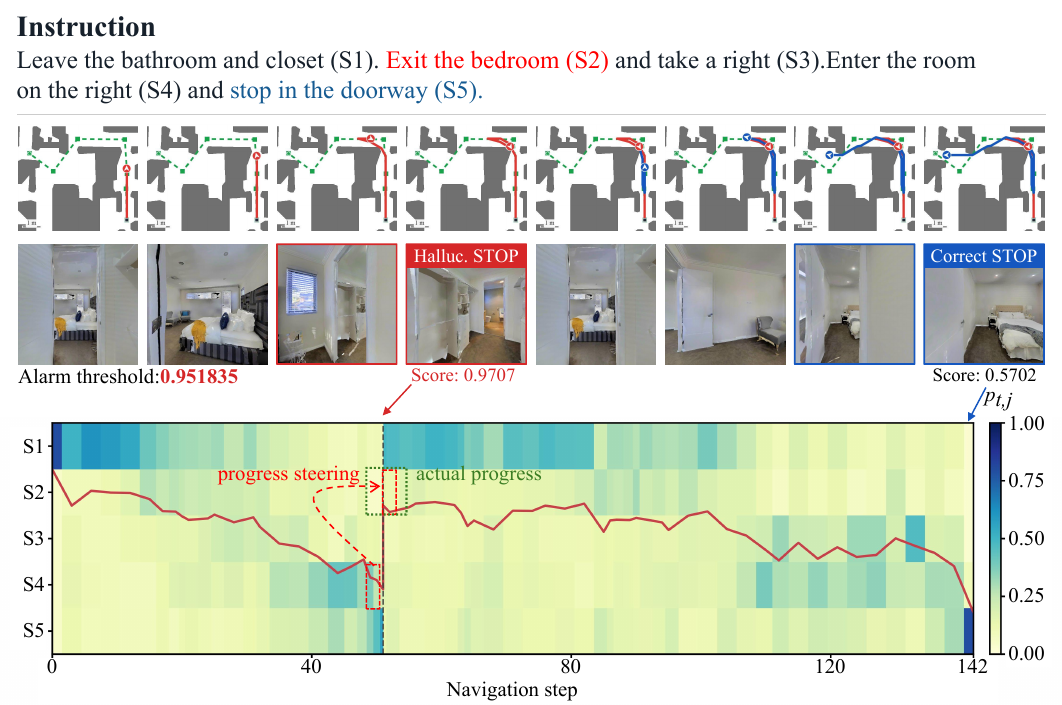}
    \caption{\textbf{Test-time Recovery with PHAR framework.} The heatmap shows sub-instruction attention $p_{t,j}$, and the red curve traces the instruction progress pointer $\mu_t$.The agent initially grounds its current observation in a later sub-instruction $S4$, advancing $\mu_t$ beyond the actual task stage. PHAR Steers $\mu_t$ toward the correct stage $S2$, enabling the agent to resume navigation and reach the goal.}
    \label{fig:vis_ana_2}
\end{figure}

Figure~\ref{fig:vis_ana_2} illustrates how PHAR detects navigation failure and steers instruction progress to resume navigation.
The agent prematurely stops in the bedroom while attending to the final sub-instruction ($S_5$), although it has not reached the specified doorway.
The verifier flags this mismatch between visual observations and instruction progress, and PHAR selects a partial history-truncation action that retains recent visual context while shifting attention back toward $S_2$, an appropriate sub-instructions consistent with the current observation.
The agent then resumes navigation, with instruction progress advancing toward later sub-instructions as it approaches and ultimately reaches the target.
This example shows how intervening on instruction progress can resume navigation after a premature \texttt{STOP} and support recovery.

\section{Conclusion}
\label{sec:conclusion}

We identify progress hallucination as a failure mode in continuous vision-and-language navigation: off-route observations can advance an agent's internal instruction progress beyond the actual task stage, leading to navigation failure.
To address this issue, we propose PHAR, a test-time self-recovery framework following a verify-then-steer paradigm.
A temporal probe with conformal calibration detects unreliable termination decisions, while contextual-bandit-based progress steering adaptively selects progress steering operators to rewind instruction progress and resume navigation.
Experiments on R2R-CE and RxR-CE demonstrate consistent improvements across multiple navigation baselines without finetuning VLMs.
These findings highlight instruction progress as a useful intervention point for enhancing the reliability of VLN agents at test time.
\section{LLM Usage Statement}
We used a large language model (LLM) to assist with code revisions and manuscript editing, including refining wording, correcting grammar, and improving readability. The authors developed the research ideas, designed the methodology, made all experimental decisions, and analyzed the results. The authors also verified the research work.
{
    \small
    \bibliographystyle{ieeenat_fullname}
    \bibliography{main}
}

\clearpage
\setcounter{page}{1}
\maketitlesupplementary
\appendix

\section*{Reference Convention}
All references in this supplementary material follow
the numbering in the main paper unless otherwise specified.

\section{Additional Investigation of Progress Hallucination}
\label{sec:supp_recovery}

The empirical study in Sec.~\ref{sec:progress_analysis} shows that the steering operator $\mathcal S_\rho$ can mitigate progress hallucination by rewinding the progress pointer toward earlier sub-instructions.
We further investigate how steering strength $\rho\in\{0.2,0.4,0.6,0.8,1.0\}$ affects navigation recovery through quantitative and qualitative analyses.

\subsection{Quantitative Analysis.}
\label{sec:supp_adaptive_evidence}

Following the experimental setup in Sec.~\ref{sec:progress_analysis}, we analyze how progress steering strength affects recovery success on those 100 failed trajectories. We apply the steering operator $\mathcal S_\rho$, with $\rho\in\{0.2,0.4,0.6,0.8,1.0\}$ independently at the same failed termination states of these trajectories. Each steering operator removes the earliest fraction $\rho$ of historical observations, after which the agent replans using the truncated history for recovery. We report the number of recovered trajectories under different steering strengths in Table~\ref{tab:supp_controlled_crop_recovery}.

\begin{table}[thbp]
    \centering
    \caption{\textbf{Navigation recovery under different steering strengths $\rho$.} Each steering operator $\mathcal S_\rho$ is applied independently to the same termination states of 100 failed trajectories. Union counts distinct trajectories recovered under at least one steering strength.}
    \label{tab:supp_controlled_crop_recovery}
    \begingroup
    \small
    \setlength{\tabcolsep}{4pt}
    \renewcommand{\arraystretch}{1.08}
    \begin{tabular}{@{}lc@{}}
        \toprule
        Steering Strength $\rho$ & \#Recovered Trajectories \\
        \midrule
        0.2  & 5 \\
        0.4  & 11 \\
        0.6  & 17 \\
        0.8  & 23 \\
        1.0 & 28 \\
        \midrule
        Union & \textbf{40} \\
        \bottomrule
    \end{tabular}
    \endgroup
\end{table}

Across the tested strengths, stronger steering improves recovery performance, with the number of recovered trajectories increasing from 5 at $\rho=0.2$ to 28 at $\rho=1.0$.
This trend is consistent with our expectation that failed trajectories may exhibit substantial progress hallucination and therefore require stronger steering of instruction progress to enable recovery.
However, the instruction progress appropriate for recovery may differ across failed trajectories.
As shown in the table, 40 of the 100 failed trajectories can recover under at least one tested steering strength, whereas only 28 trajectories successfully recover under the best fixed strength. This highlights the need to select appropriate steering strengths for different failed trajectories to maximize the recovery rate.

\subsection{Qualitative Analysis.}
\label{sec:supp_qualitative}

\begin{figure*}[thbp]
    \centering
    \includegraphics[width=0.9\textwidth]{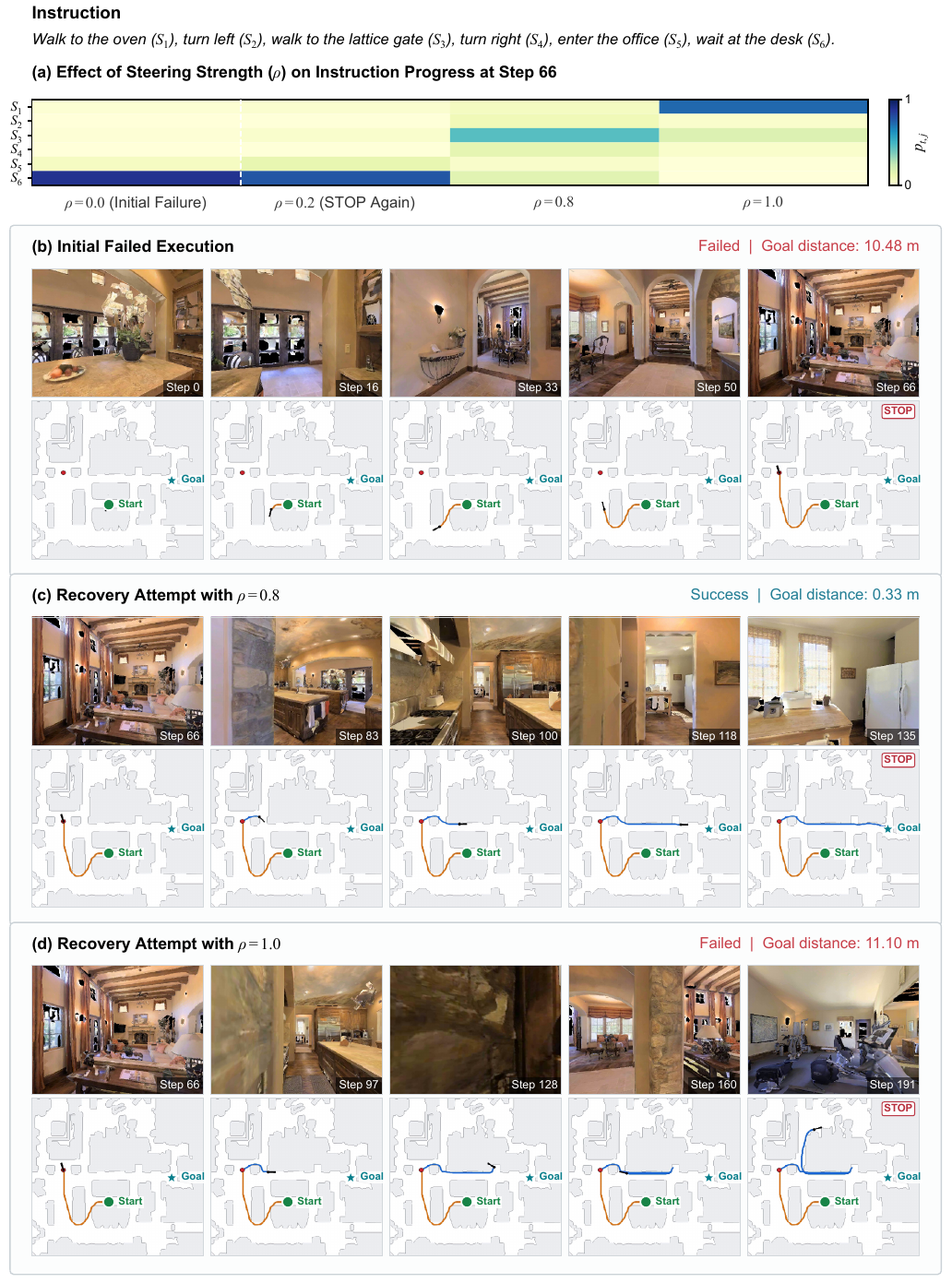}
    \caption{\textbf{Recovery requires an appropriate steering strength.}
    (a) Instruction progress distribution $p_t$ (defined in Sec.~\ref{sec:progress_analysis}) under different steering strengths: $\rho=0.2$ triggers another \texttt{STOP}, $\rho=0.8$ redirects attention toward $S_3$ and enables recovery, while $\rho=1.0$ shifts attention back to $S_1$ and leads to unsuccessful exploration.
    (b--d) Sampled observations and maps show the initial failed execution and recovery attempts with different steering strengths $\rho$. Orange and blue paths indicate the original and recovery trajectories.}
    \label{fig:supp_strength_case}
\end{figure*}

To understand why different steering strengths lead to different recovery outcomes, we examine a failed NaVIDA trajectory from the R2R-CE val-unseen split (Fig.~\ref{fig:supp_strength_case}b).
The instruction is \textit{``Walk to the oven $(S_1)$, turn left $(S_2)$, walk to the lattice gate $(S_3)$, turn right $(S_4)$, enter the office $(S_5)$, wait at the desk $(S_6)$.''}
Ambiguity in grounding the instruction's landmarks can lead to incorrect route choices.
In this case, the agent deviates from the intended route and terminates before reaching the office and target desk. Nevertheless, its attention is concentrated on the final waiting instruction $S_6$ (Fig.~\ref{fig:supp_strength_case}a), indicating progress hallucination: the agent treats off-route observations as evidence of task progress and predicts \texttt{STOP} prematurely.

We independently apply the progress steering operator $\mathcal S_\rho$ with three different strengths $\rho\in\{0.2,0.8,1.0\}$ at this failed termination state.
With $\rho=0.2$, attention remains concentrated on $S_6$, indicating that progress hallucination persists, and the agent immediately predicts \texttt{STOP} again (Fig.~\ref{fig:supp_strength_case}a).
At $\rho=0.8$, attention shifts toward $S_3$, which instructs the agent to walk to the lattice gate.
The agent reorients to seek this intermediate landmark, then continues toward the office and successfully reaches the target desk (Fig.~\ref{fig:supp_strength_case}c).
At $\rho=1.0$, however, attention shifts back to the initial oven-related instruction $S_1$.
Although navigation resumes, the agent engages in unnecessary exploration and ultimately stops at an incorrect location (Fig.~\ref{fig:supp_strength_case}d).
These contrasting outcomes suggest that effective recovery requires steering instruction progress toward an appropriate stage: insufficient steering leaves progress hallucination unresolved and leads to another premature \texttt{STOP}, whereas excessive steering shifts attention to overly early sub-instructions and prompts unnecessary exploration that can hinder recovery.
This motivates our Adaptive Progress Steering Module (Sec.~\ref{sec:steering}), which selects appropriate steering strengths for online recovery.

\section{Real-World Experiments}
\label{sec:supp_realworld}

We evaluate our framework in a real-world setting to examine whether PHAR can support recovery from actual navigation failures and improve the reliability of real-world navigation systems.

\subsection{Deployment Setup}
We deploy PHAR with NaVILA~\cite{cheng2024navila} on a Unitree Go2 EDU quadruped robot using a cloud-assisted architecture.
The robot is equipped with an RGB camera, and NaVILA uses only monocular RGB observations for navigation.
The robot's onboard NVIDIA Jetson Orin NX runs the low-level controller, while a remote server with a single NVIDIA RTX 4090 GPU hosts NaVILA and PHAR's verification and progress steering components.
The robot periodically sends RGB observations to the server via HTTP.
The server performs navigation inference and returns textual action commands, which the onboard controller converts into motor commands.
When the verifier flags a proposed \texttt{STOP} as unreliable, adaptive progress steering rewinds the agent's instruction progress, prompting NaVILA to recover from navigation failure based on earlier sub-instructions.

\subsection{Results}
We present an indoor navigation failure case and demonstrate recovery using PHAR.
The task instruction is \textit{``Walk past the plants and continue until you reach the bookshelf $(S_1)$. Turn right at the bookshelf $(S_2)$, then proceed to the green chair and stop there $(S_3)$.''}
For this task, we construct a challenging environment with two identical green chairs: one visible ahead of the robot and the intended target initially occluded to the right of the bookshelf. This arrangement introduces ambiguity in grounding the instruction under partial observability, making the agent susceptible to mistaking the visible chair for the target and overestimating its instruction progress. 

\begin{figure*}[htbp]
    \centering
    \includegraphics[width=0.9\textwidth]{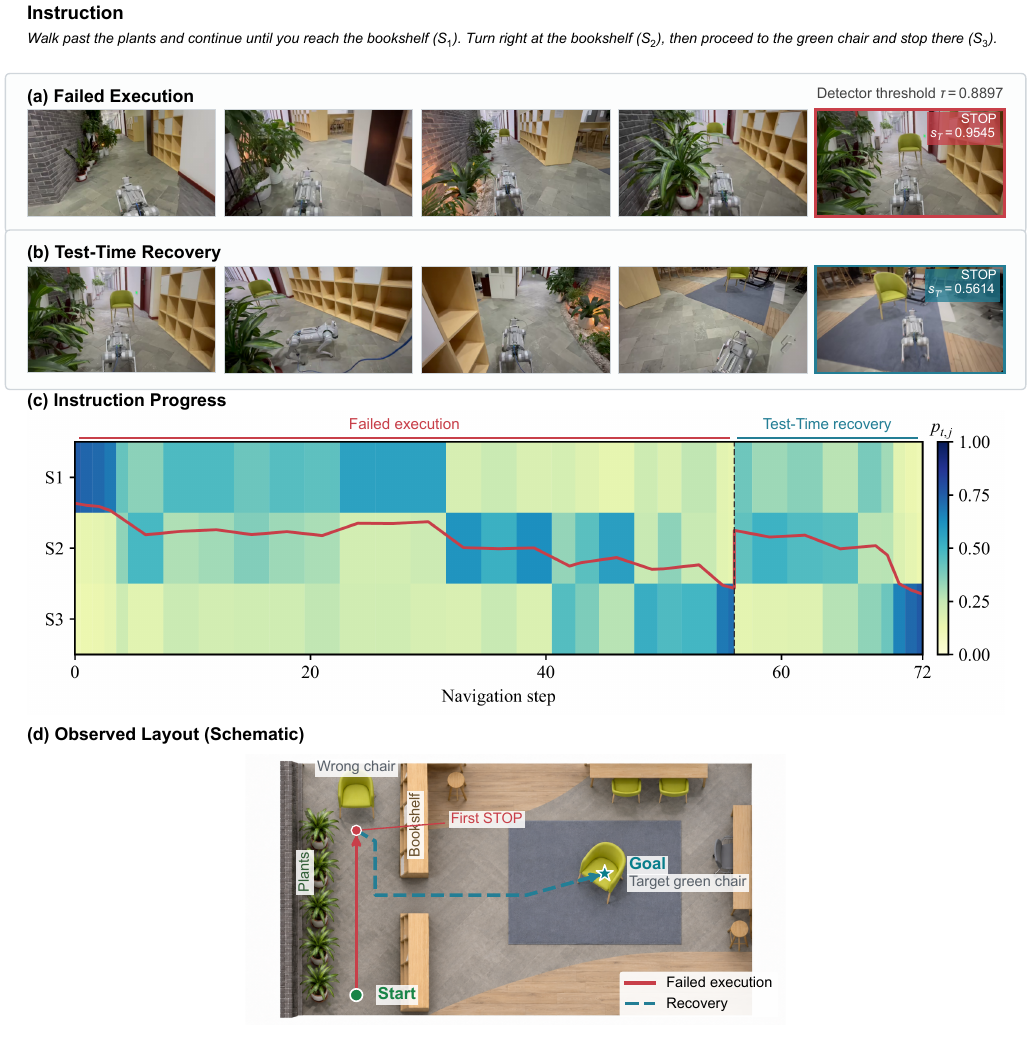}
    \caption{\textbf{Real-world recovery with PHAR.}
    (a) NaVILA prematurely stops near the wrong green chair; our verifier detects the unreliable termination ($s_T=0.9545$).
    (b) Our Adaptive Progress Steering Module redirects instruction progress, enabling the robot to complete the missed right turn and reach the target chair, where termination is accepted ($s_{T'}=0.5614$).
    (c) Instruction progress distribution $p_t$ and pointer $\mu_t$ (red curve; see Sec.~\ref{sec:progress_analysis}) illustrate the effect of steering; the dashed line marks the intervention.
    (d) Schematic layout showing the two chairs, the passage between bookshelf segments, start and goal, and illustrative failed (red) and recovery (blue) routes.}
    \label{fig:supp_realworld_case1}
\end{figure*}

As shown in Fig.~\ref{fig:supp_realworld_case1}, the agent fails to observe the target chair in the occluded area and continues straight, proposing \texttt{STOP} near the wrong green chair before completing the instructed turn. The observations collected along this off-route segment lead the agent to overestimate its instruction progress. Analysis of the model's internal state reveals that its instruction progress advances beyond the actual task stage: despite not completing $S_2$ (\textit{``turn right at the bookshelf''}), the agent anchors its attention on the final sub-instruction $S_3$, exhibiting progress hallucination. This makes it difficult for the baseline agent to attend to earlier sub-instructions and correct the navigation failure.

With PHAR, the verifier detects the erroneous termination caused by progress hallucination, assigning it a failure score of $s_T=0.9545$ and flagging it as unreliable.
Our Adaptive Progress Steering Module then steers the agent's instruction progress backward, redirecting its attention to sub-instructions $S_1$ and $S_2$.
Guided by these earlier sub-instructions, the robot resumes navigation, completes the turn at the bookshelf, and reaches the target green chair.
The verifier score at this second termination decreases to $s_{T'}=0.5614$, and the termination passes verification.
This case illustrates how PHAR combines verification and progress steering to recover from navigation failure in a physical environment.

\end{document}